# 14. Artificial Intelligence (AI) in Legal Data Mining

Aniket Deroy, Naksatra Kumar Bailung, Kripabandhu Ghosh, Saptarshi Ghosh and Abhijnan Chakraborty

*"Artificial intelligence is growing up fast, as are robots whose facial expressions can elicit empathy and make your mirror neurons quiver."*
- Diane Ackerman, American poet

Summary:
- Despite the availability of vast amounts of data, legal data is often unstructured, making it difficult even for law practitioners to ingest and comprehend the same.
- It is important to organise the legal information in a way that is useful for practitioners and downstream automation tasks.
- The word 'ontology' was used by Greek philosophers to discuss concepts of existence, being, becoming and reality. Today, scientists use this term to describe the relation between concepts, data, and entities.
- A great example for a working ontology was developed by Dhani and Bhatt. This ontology deals with Indian court cases on intellectual property rights (IPR)
- The future of legal ontologies is likely to be handled by computer experts and legal experts alike.

## 1. Introduction

The history of the use of Artificial Intelligence (AI) in legal data mining goes back to the time when legal experts began collaborating with computer scientists to solve compelling practical problems in law. With an enormous amount of legal data becoming available across jurisdictions, algorithms to make sense of the same became a necessity. The broad domain of AI-based legal data mining gained prominence through a sub-field called E-discovery prevalent in the courts of the United States. "Electronic Discovery" or "E-Discovery"[1] refers to the phenomenon of producing information stored in hard drives, network servers and emails as evidence in a litigation. In 2009, the Sedona Conference, a leading conference in law, recognised that "the legal profession is at a crossroads", and openly expressed willingness of the fraternity to embrace AI, at least in E-discovery. The associated legal body requested industry and academia to engage with them to devise AI-based solutions for several legal problems. This led to the establishment of Legal Track in the popular Text Retrieval Conference (TREC), which seeks to find AI-based solutions to E-discovery problems by simulating real-life legal scenarios. Consequently, in collaboration with legal professionals, AI researchers developed algorithms on practical problems in the field. India followed suit, with similar endeavours in the form of legal tracks in the conference Forum for Information Retrieval Evaluation (FIRE). The researchers created testbeds for real-life legal problems such as prior case retrieval and statute retrieval, legal document summarisation and so on.

Although several legal tasks have benefitted from the application of AI algorithms, in this chapter, we focus on two specific tasks -- (i) rhetorical role labelling and (ii) summarisation of legal documents. The first task involves the labelling of every sentence in a case document (e.g., Facts, Arguments, Issues, Judgement), representing the semantic function the sentence serves. We discuss multiple state-of-the-art AI algorithms proposed for rhetorical role labelling. The second task is relevant for legal practitioners who need to check hundreds of case documents. There have been

---

[1] **References**

Oard, D.W., Baron, J.R., Hedin, B., Lewis, D.D. and Tomlinson, S., 2010. Evaluation of information retrieval for E-discovery. *Artificial Intelligence and Law, 18*(4), pp.347-386.

multiple developments in text summarisation methods. In this chapter, we cover both domain-specific summarisation methods that utilise the nuances of legal domain as well as domain-independent summarisation algorithms that do not explicitly focus on the nitty-gritty of legal documents.

Despite the availability of vast amounts of data, legal data is often unstructured, making it difficult even for law practitioners to ingest and comprehend the same. It is therefore important to organise the legal information in a way that is useful for both practitioners as well as the downstream automation tasks. Towards this goal, several approaches have been considered. In this chapter, we specifically focus on legal knowledge bases as a useful approach for data management. Historically, legal knowledge bases have been generated manually by legal experts who inculcated rules to serve specific use-cases. Recently, many frameworks have been proposed to automate the task, including LEGIS (for criminal cases), JurWordnet (Wordnet on legal documents), Law Article Ontology (for intellectual property rights and licensing of digital content) and DALOS ontology (for consumer protection). In the next section, we dive deep into these exciting techniques, and discuss the advantages and pitfalls of different approaches. The rest of this chapter is meant to be a technical review of the application of AI in the important area of legal analytics. Wherever required, references are provided to basics of the specific concept / technique in question. The reader is advised to benefit from this chapter accordingly.

## 2. Legal Knowledge Base and Ontologies

The word 'ontology' was used by the Greek philosophers to discuss concepts of existence, being, becoming and reality. Today, scientists use this term to describe the relation between concepts, data and entities that substantiate one or multiple domains of discourse. Specifically, ontology is used to describe things or phenomena in the domain of interest, leading to the construction of a knowledge graph. It serves as the vocabulary of the domain, indicating terms that can be used and the semantics and sentences that can be expressed. Originally, the primary purpose of ontology was to offer semantics for the Semantic Web – the definitions and relationships between the terms that make up a given domain. These semantics are usually expressed in the Web Ontology Language (OWL), a knowledge representation formalism that supports machine-based inference and reasoning. Figure 1 shows an example ontology representing the relationship between different work roles and the surrounding context.

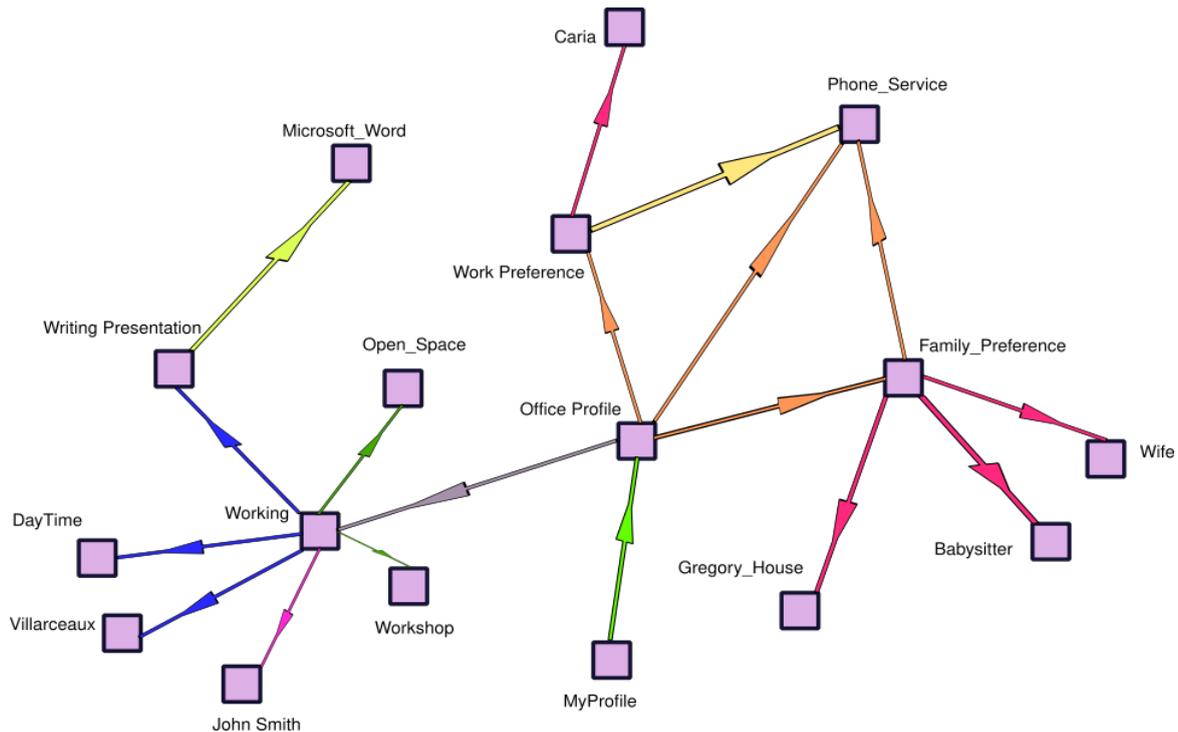

Figure 1: An example ontology to infer a person's working tools and situation
(Figure reproduced from[2])

Similarly, legal ontologies describe the universe of discourse in the legal domain. It can be useful to academics, laypeople, and attorneys in a variety of situations, including modelling legal proceedings, semantic search and indexing, and keeping up with the constantly changing rules and regulations.

**2.1 Need for legal ontologies**
There are several uses of legal ontologies, listed as follows:

A. **Semantic indexing and searching**
Ontologies are often used as query-based systems to represent document content and other related data. In the legal domain, ontologies help legal professionals and researchers deal with knowledge in the form of documents, diagrams, schemas, etc. In situations of a single word with multiple meanings, the ontologies need to deal with these words individually. In such cases, manual annotations can further help the ontologies for application in a more targeted manner.

B. **Reasoning and problem-solving**
The basic role of ontology is to express the knowledge of a particular domain, enabling reuse of generic knowledge. It is possible to develop knowledge bases that not only address the issue at hand but are also more manageable and extensible. Such ontologies aid inference.

C. **Understanding the domain**

---

An ontology can be viewed as a map of knowledge that can be identified in the domain. These are aptly named core ontologies because they represent the domain and aid the acquisition of relevant knowledge.

D. **Semantic integration and interoperation**

Ontologies enable the interoperation of different information systems. They define the vocabulary for the interchange of information, functioning as a semantic information schema. Such ontologies usually reuse parts of ontologies created for other uses.

An ontology captures the meaning of terms. However, the meaning cannot be specified independent of the context of use. Consider the word "house"; it could mean a commodity (for the owner), an assembly (for the government) or even gambling establishments. Thus, a concept's meaning is tied to the specific sense it expresses. Top-level ontologies represent concepts such as time, location, causality, object, process, etc. that should be constant among ontologies. A core-ontology contains the ideas that are constant across a certain field of practice. For example, a core law ontology will typically encompass concepts such as obligation and right and responsibility, which can be specialisations of expectation, desirability, intention, and causality.

## 2.2 Examples of legal ontologies

With the premier International Conference on Artificial Intelligence and Law (ICAIL) nearing 35 years of establishment, many legal ontologies are now part of the literature. Although ontologies can be both single as well as multi-domain, there is a clear preference for the single domain. Some of the popular single-domain ontologies are listed below:

I. LegalRuleML3 (LegalRule Markup Language)[3]: It is based on interpretations of a rule, tracking the author of a document and models of temporal evolution of norms.
II. CC7 (Creative Common Rights Expression Language)[4]: It is based on concepts from Creative Commons licence.
III. L4LOD8 ((Licences for Linked Open Data)[5]: It is based on the actions applicable as well as disallowed on Linked Open Data.
IV. GDPRtEXT11 (General Data Protection Regulation)[6]: It models the concepts expressed in GDPR as Linked Data, along with the structure of the GDPR text.

## 2.3 Core ontologies

The ontologies mentioned above are an amalgamation of core ontologies based on specific legal texts. In philosophical principles, a core ontology is a basic and minimal ontology containing only the base concepts required to comprehend other concepts. In information science, core ontology contains general terms common across the domain. While there are arguments about the feasibility of such domains, these are the best models (in theory) to capture the logical form and metaphysics of the labels across the domains. Next, we discuss two popular core ontologies used in legal knowledge bases: FOLaw and LRI-Core.

---

[3] *Standard, O.A.S.I.S., 2021. LegalRuleML Core Specification Version 1.0.*
[4] Abelson, H., Adida, B., Linksvayer, M. and Yergler, N., 2008. ccREL : The Creative Commons Rights Expression Language.
[5] Governatori, G., Rotolo, A., Villata, S. and Gandon, F., 2013, October. One license to compose them all. In *International semantic web conference* (pp. 151-166). Springer, Berlin, Heidelberg.
[6] Pandit, H.J., Fatema, K., O'Sullivan, D. and Lewis, D., 2018, June. GDPRtEXT-GDPR as a linked data resource. In *European Semantic Web Conference* (pp. 481-495). Springer, Cham.

### 2.3.1 FOLaw

In contrast to most legal theoretical studies, FOLaw (Functional Ontology for Law)[7] presents a legal-sociological viewpoint. There is also a parallel idea of "functional" at play. The roles that different "types of knowledge" play in the reasoning are captured by FOLaw, which identifies the dependencies between them.[8]

FOLaw has a variety of uses. The first one focuses on identifying the main categories of knowledge used in legal reasoning and outlining how they are interdependent. An architecture for legal reasoning could be simply translated from this typing and its dependencies. The second function is a typical core-ontology role: It entails cataloguing and organising collections of domain ontologies as well as assisting in the knowledge acquisition for new ontologies. FOLaw describes the following six types of knowledge:

I. **Normative knowledge**

The basic conception of norm expresses it as an idealisation that should be the case or occurs as per the will of the agent that created the norm. Models and roles are also idealisations with respect to reality. One can play the role according to these "guidelines" or fail, similar to the situation where one violates a norm.

II. **Meta-legal knowledge**

Meta-legal knowledge is leveraged for resolving conflicts between norms and to identify which legal knowledge is valid.

III. **World knowledge**

The relationship between the legal system and actual events in jurisdictions is mediated and filtered by global knowledge.

IV. **Responsibility knowledge**

Responsibility knowledge establishes or disestablishes a connection between a violation of a norm and an agent who is to be held accountable (guilty, liable) for this violation. It assigns or restricts the responsibility of an agent over a specific (disallowed) situation.

V. **Reactive knowledge**

It outlines the reaction that ought to be taken in a given context and how. Normally, this response is a punishment, but it might even be a "prize" (for example, government benefits).

VI. **Creative knowledge**

Since creative knowledge is independent of other kinds of knowledge, it is relatively isolated in FOLaw. Using what we refer to as "creative knowledge", legislation may permit the development of new or fictitious legal entities, typically some organisation or other legal person.

---

[7]      Breuker, J., 2003. The construction and use of ontologies of criminal law in the ecourt european project. *Proceedings of Means of electronic communication in court administration*, pp.15-40.

[8]      Breuker, J. and Hoekstra, R., Epistemology and ontology in core ontologies: FOLaw and LRI-Core, two core ontologies for law. In *Proceedings of EKAW Workshop on Core ontologies [Internet]. Northamptonshire, UK: Sun SITE Central Europe.*

### 2.3.2 LRI-Core

LRI-Core is a more generic core ontology developed to overcome the epistemological promiscuity of FOLaw. LRI-Core was created using FOLaw rather than being based on any top or foundational ontologies that already exist. The top level of the legal core ontology is divided into five components, or worlds: mental concepts, physical concepts (object and process), abstract concepts, roles, and occurrences.

**I. Physical world**

The physical world contains two main classes: processes and physical objects. Anything with matter, extension, shape, or aggregate state qualifies as an object. Processes are the modifiers of objects and the links between them.

In LRI-Core, processes are classified based on two perspectives: (1) The formal kind of change (transformation, transduction, and transfer), and (2) the characteristics of involved objects, such as movements altering position or chemical process altering substance.

**II. Mental world**

Concepts are the building blocks of mental objects such as thoughts, and mental objects' substance are representations. Propositional attitudes, such as belief, desire, and norm, form the conceptual content of thoughts.

**III. Roles**

Roles cover functional views on mental processes, on agent behaviour or on physical objects. Roles by definition are idealisations that do not exist. An important distinction should be noted between playing a role and the role itself: "agents can act, and roles cannot". Social roles are sometimes complementary (student-teacher, speaker-hearer). Regarding the reciprocity of legal situations, or the rights and obligations that go with them, there is a related theory of law. Roles can be defined as behavioural requirements on role execution or prescriptions that are formally enforced by law. When our behaviour departs from the norms associated with these roles, we break the law.

**IV. Abstractions**

One could counter that all concepts are abstractions, therefore impossible to see a separate abstract world. But it is important to note that a (relatively) small number of proto-mathematical ideas, such as collections, sequences, and countable positive numbers, are thought of as common sense. Since semi-formal abstractions have a minor role in legal texts, LRI-Core has only a few abstract classes.

**V. Occurrences**

Ontologies provide the classes with which we can identify individual entities in situations. Only the classes of the entities that make up a situation are included in an ontology; instances (occurrences) themselves are not. A separate framework can be used to capture occurrences.

**A real-world application of LRI-Core**

The e-court European IST project[9] has an ontology that was formalized in the Web Ontology Language-Description Logic (OWL-DL)[10] using Protégé and was formalized in LRICore as Resource Description Framework (RDF(S)) with Protégé. RDF is the standardized designed data model for metadata for objects. RDFs can link objects using relations and describe the two ends of a relationship, which is a convenient method for graph data exchange. LRI-Core had a total of around 100 concepts. It served as a tool for the e-court project's construction of ontologies, notably the Dutch criminal law ontology, which supported knowledge acquisition. The authors nevertheless concluded that "the number of legal concepts in LRI-Core was rather small; it was rather a top ontology covering abstract concepts of common-sense rather than the field of law"[11].

## 2.4 Ontologies in practice

We have seen a few foundational ontologies, but these are typically the ontologies that actual ontologies are created from. Core ontology development is more of a design process than a data-driven procedure. Ontologies are consequently trained on particular data or on particular context in practice. We then go over a particular practical ontology.

### 2.4.1 LegalRule Markup Language (LegalRuleML)

RuleML has been expanded into the legal industry as LegalRuleML. Rules are used to represent procedures, rules, policies, logic, and declarative programmes that should be used in these circumstances. Rules are defined as the interactions between causes and effects (also known as "laws"), situations and actions (also known as "triggers"), and premises and implications (also known as "implications"). To build up to a broader category of rules in a hierarchy, a subset of these straightforward rules is contained in a RuleML document.

The lack of a standardised language to define rules that were generated or utilised inside of ontologies, with translators in and out as well as other tools, presented a challenge to RuleML designers. When different legal ontologists wanted to exchange their efforts, the Relational-Functional Markup Language (RFML), UML-Based Rule Modelling Language (URML), and Agent Object Relationship Markup Language (AORML) caused disagreements. To create an open, vendor-neutral XML/RDF-based rule language, this called required a concerted effort from numerous expert teams from various nations.

The Organization for the Advancement of Structured Information Standards (OASIS) has developed LegalRuleML, an extension of RuleML that adds characteristics relevant to the legal industry. By converting the compliance requirements into a machine-readable format, it can be used to model different laws, rules, and regulations by bridging the gap between natural language descriptions and semantic standards.

LegalRuleML adhered to the following fundamental design principles:

---

[9] Breuker, J., 2003. The construction and use of ontologies of criminal law in the ecourt european project. *Proceedings of Means of electronic communication in court administration*, pp.15-40.
[10] Prat, N., Megdiche, I. and Akoka, J., 2012, November. Multidimensional models meet the semantic web: defining and reasoning on OWL-DL ontologies for OLAP. In *Proceedings of the fifteenth international workshop on Data warehousing and OLAP* (pp. 17-24).
[11] Breuker, J. and Hoekstra, R., Epistemology and ontology in core ontologies: FOLaw and LRI-Core, two core ontologies for law. In *Proceedings of EKAW Workshop on Core ontologies [Internet]. Northamptonshire, UK: Sun SITE Central Europe*.

I.   **Multiple semantic annotations**: Multiple semantic annotations that represent various legal interpretations may be present on a legal rule. These are displayed as internal or external metadata in a different annotation section[12]. The interpretation is based on several factors, including provenance, the relevant jurisdiction, the logical interpretation of the rule, and others.
II.  **Tracking the creators**: Any LegalRuleML document, or any of its pieces, may contain provenance information that links it to its authors. By doing this, it was always apparent who to trust and whose knowledge base the annotations belonged to.
III. **Linking rules and provisions**: Multiple linkages between the rules and the textual provisions are possible using LegalRuleML. A provision can link to several rules, and a rule can link to many provisions.
IV.  **Temporal management**: Provisions, rules, applications of rules, references to text, and references to actual physical objects are only a few of the items that can be found in legal texts. A LegalRuleML rule has both temporal and non-temporal features, including the internal constituency of the activity, the moment of the rule's assertion, the efficacy, and enforcement (status, validity, jurisdiction).
V.   **Formal ontology reference**: LegalRuleML has a system for pointing to reusable classes of a specified external ontology but being autonomous from any legal ontology and logic framework.

**2.4.2 Use cases of LegalRuleML**

The syntax of LegalRuleML is almost identical to that of the well-known web programming languages HTML or XML. The precedence of the classes it links to is indicated by the hierarchy of angle brackets. Further, LegalRuleML likewise employs a straightforward "if-else" strategy. It looks for formulas and deontic formulas. It recommends the equations as formulae if it locates an appropriate section of text in the case containing formulas.

Some use cases of LegalRuleML are:
- LegalRuleML is used in the eHealth industry to model security and privacy concerns for controlling document access based on the operator's profile and authorizations. Using LegalRuleML, it is possible to construct several views of the same health record or document depending on the role of the querying agent and filter sensitive data in accordance with the law or regulation.

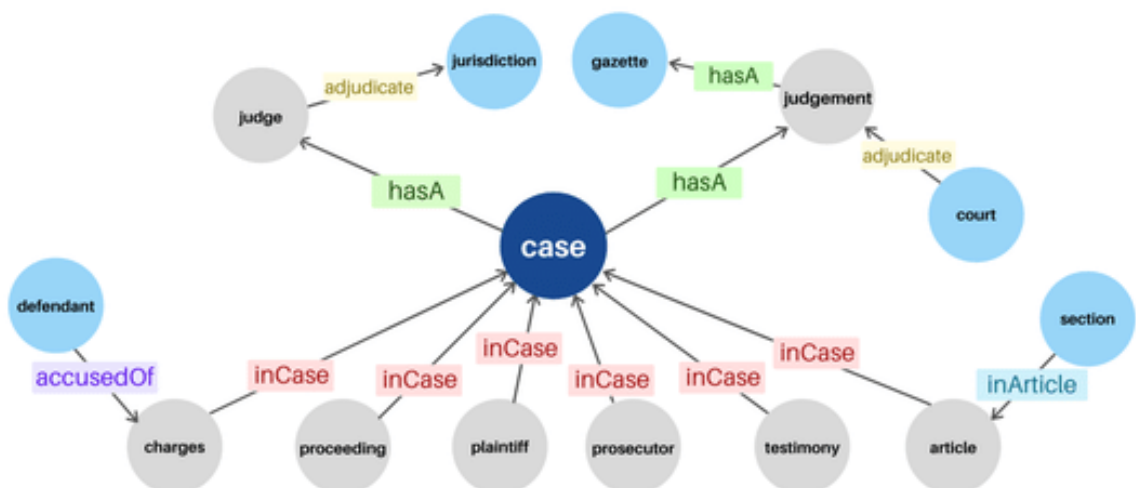

---

[12]   *Standard, O.A.S.I.S., 2021. LegalRuleML Core Specification Version 1.0.*

Figure 2: An ontology for legal documents (Reproduced from[13])

- To enable automatic IPR compatibility checks between various datasets, LegalRuleML is used in the open data domain to model the creative commons licences of the datasets. This is especially useful for determining if various datasets can be integrated to create a commercial application.
- LegalRuleML is used by MIREL, which is supported by a Horizon 2020 funding from the European Union[14]. To construct tools for MIne and REason with legal documents, a formal framework must be established. They will be transformed into formal notions that may be applied to decision-making, compliance verification, and norm querying.
- Akoma Ntoso, another comparable XML-based format created by the OASIS LegalRuleML team for judicial, legislative, and parliamentary documents, won the UNGA challenge, where the major objective was to extract knowledge from the UN General Assembly resolutions. It is currently a part of the OASIS standard.

**2.5 Using ontology for recommending similar cases**
A great example for a working ontology is developed by Dhani and Bhatt[15]. This ontology deals with Indian Court Cases on Intellectual Property Rights (IPR). They used Graph Neural Networks for creating the ontology suited for recommending similar cases. The data was taken from repositories like Indian Kanoon and Casemine. Additionally, public NLP libraries such as IBM's Data Discovery and Stanza were used to retrieve information out of the data. The authors then used a combination of manual methods and Latent Dirichlet Allocation (LDA) to obtain relevant topics.

The authors computed node similarity by taking the nodes as documents and using 'lawpoints' as additional manual features. These 'lawpoints' were leveraged to annotate IPR relevant legislations using a label from a selected number of broader and prominent IPR concepts based on the semantic similarities. They extracted topics to compute the relevance of the results obtained by a combination of Term frequency and Inverse Document Frequency (TF-IDF). By ranking these words, some of the concepts related to IPR such as 'patent' and 'copyright' emerged from the data organically which were then selected as 'lawpoints'.

---

[13]    Dhani, J.S., Bhatt, R., Ganesan, B., Sirohi, P. and Bhatnagar, V., 2021. Similar cases recommendation using legal knowledge graphs. *arXiv preprint arXiv:2107.04771*.
[14]    Shandilya, A., Dash, A., Chakraborty, A., Ghosh, K. and Ghosh, S., 2020, December. Fairness for Whom? Understanding the Reader's Perception of Fairness in Text Summarization. In *2020 IEEE International Conference on Big Data (Big Data)* (pp. 3692-3701). IEEE.
[15]    Dhani, J.S., Bhatt, R., Ganesan, B., Sirohi, P. and Bhatnagar, V., 2021. Similar cases recommendation using legal knowledge graphs. *arXiv preprint arXiv:2107.04771*.

Figure 3: An ontology for legal documents (Accessed from[16])

The ontology used citations and similarity between cases to be possible relations among the cases and was created using Relational Graph Convolution Networks[17] to predict the labels in the documents. AUC scores were used to check the performance of the ontology thus created, which returned a score of 0.620 for the citation graph and 0.556 for the similarity. There was a significant decrease of score (almost 0.04) when the ontology did not use 'lawpoints' as features.

We can see in figure 3, this ontology makes use of 14 main features and 5 relationships are shown here. Using TF-IDF and topic modelling, the authors also got relevant words from the data, as seen in the table of figure 4. Some manual categorizing of these features was done to create 'lawpoints' as discussed above, among which some of the main ones were from IPR concepts ('patent', 'copyright') and some extracted from the document ('section', 'court', 'plaintiff'). These extracted topics exemplified the domain specificity of the ontology being expanded upon and the relations that the entities shared with each other.

**2.6 Challenges in developing legal ontologies**
While developing technologies for legal domain, there are multiple challenges[18]; some of them are listed below:
  I. Ontologies in general require merging of multiple concepts which sometimes leads to not-so-great results.

---

[16] Dhani, J.S., Bhatt, R., Ganesan, B., Sirohi, P. and Bhatnagar, V., 2021. Similar cases recommendation using legal knowledge graphs. *arXiv preprint arXiv:2107.04771.*

[17] Schlichtkrull, M., Kipf, T.N., Bloem, P., Berg, R.V.D., Titov, I. and Welling, M., 2018, June. Modeling relational data with graph convolutional networks. In *European semantic web conference* (pp. 593-607). Springer, Cham.

[18] Ni, Zhang & Wang, Ping & PU, Yi-Fei. (2015). Challenges and Related Issues for Building Chinese Legal Ontology. 10.2991/meic-15.2015.287

II. Legal ontology retrieval is an analogue for human thinking and logical reasoning is the bridge for statutes and judicial precedents. Hence the logic entailed in legal ontologies is usually never complete, and even does not include innovation.
III. Legal concepts appear explicit, but the contents are large; therefore, controlling factors inside it is harder than expected.
IV. Law is dependent upon the time, politics, culture, and many other factors prevalent in a country at a period. Training ontologies while using these parameters might lead to an immense compilation time and not considering them only creates ontologies which in theory does not encompass all the factors happening around the time.

## 2.7 Future of legal ontologies

With increasing use of technologies, the future of legal ontologies is likely to be handled by computer experts and legal experts alike. One must remember that in the legal domain, ontologies must serve the correct purpose and hence provide correct answers based on the queries provided to it by users. The relevance of the ontology will be a huge factor as laws change quite frequently.

With the development of powerful and user-friendly interfaces, accessibility of ontologies has a bright future ahead of itself. The clarity of the data provided into the ontologies will be an important factor for the quality of the generated ontology. Thus, most of the pre-computer era legal documents need to be checked for errors and omissions. We hope that as new legal data is generated, better ontologies will emerge with time.

## 3. Rhetorical Role Classification

After discussing legal ontologies for efficient storage of legal data, next we turn towards two downstream tasks: Rhetorical Role Classification and Legal Document Summarization.

### 3.1 What are Rhetorical Roles?

Rhetorical Role[19] of a particular sentence in a legal document refers to the semantic function/ meaning that the sentence carries in the document. Understanding the rhetorical roles of different sentences allows us to understand the document in a modular fashion. Traditionally, rhetorical role labelling has been performed manually by legal experts. Recently, different AI based computational approaches have been developed for automatic identification of rhetorical roles. However, even for the AI techniques to work, legal expertise is still necessary to annotate an initial set of legal documents with rhetorical roles, which the AI algorithm can take as training data and learn to perform the task without human supervision.

Researchers have tried rhetorical role labelling for legal documents from different countries. Farzindar et al.[20] used four rhetorical markers to divide legal documents from the US Supreme Court into segments namely Introduction, Context, Judicial Analysis and Conclusion. Similar methods have also been attempted on Indian legal documents for segmenting an entire document into meaningful units, which identify the case, establish the facts, arguments, history of the case, and final judgement. More specifically, Bhattacharya et al[21] considered the following 8 rhetorical roles:

I. **Facts -** These are the descriptions of events that have led to the filing of a legal case. Note that these events are not still proved or unproved in the court of law based on the legal principles of the corresponding justice system.
II. **Argument** - This refers to the arguments which the lawyers of the defending parties present in-front of the judge to represent their side of the case and question the opposite side.
III. **Issues -** This refers to the legal questions whose answers are being sought, and which must be decided by the judge.
IV. **Ratio of the decision** - This refers to the reasons provided by the judge(s) to explain the final judgement of the case.
V. **Ruling by lower court** - In the Indian Judiciary, the judgement given by a particular court is not final and can be challenged in a higher court or a higher system of justice as suitable. So, a challenger case document also includes the judgements given by a lower court.
VI. **Ruling by present court** - This refers to the judgement given by the judge of the court where the case hearing took place.
VII. **Precedents** - This rhetorical role refers to the past legal cases which act as guidance for future legal cases of the similar type.
VIII. **Statutes** - This refers to the acts, notifications, orders, sections, articles, etc. which directly influence the final decision of the case. Statutes act as strong foundations based on which final decisions could be given on a legal case.

---

[19] Deroy, A., Bhattacharya, P., Ghosh, K. and Ghosh, S., 2021. An Analytical Study of Algorithmic and Expert Summaries of Legal Cases. In *Legal Knowledge and Information Systems* (pp. 90-99). IOS Press.

[20] Farzindar, A. and Lapalme, G., 2004. LetSum, an automatic Legal Text Summarizing System. In *the Seventeenth Annual Conference* on *Legal Knowledge and Information Systems* (Vol. 120, p. 11). IOS Press.

Table 1: Examples of rhetorical roles in a case document [Reproduced from web source]

| One Dwarka Nath Ghose was the owner of considerable moveable and immoveable properties. | Fact |
|---|---|
| It was contended on behalf of the appellants that the dedication of the premises. | Argument |
| The appeal is concerned only with the premises. | Issue |
| The first contention of the appellants is clearly untenable on the very language of the will of Dwarka Nath. | Ratio of the decision |
| The said suit was heard by Justice Bose who declared the premises. | Ruling by Lower Court |
| The first contention of the appellants therefore fails, and we hold that the dedication of the premises. | Ruling by Present Court |
| rankin c j in surendrakrishna ray v shree shree ishwar bhubaneshwari thakuran 1933. | Precedent |
| The prevention of corruption act 1947 by major somnath accused. | Statute |

---

[21] Ghosh, S. and Wyner, A., 2019, December. Identification of rhetorical roles of sentences in Indian legal judgments. In *Legal Knowledge and Information Systems: JURIX 2019: The Thirty-second Annual Conference* (Vol. 322, p. 3). IOS Press.

## 3.2 How to automatically detect Rhetorical Roles?

Automatically identifying the rhetorical role of a sentence is a challenging task. Researchers have used a multitude of machine learning models to correctly classify sentences into rhetorical roles.

### 3.2.1 Conditional Random Fields

Conditional Random Fields (CRF)[22][23][24] can be used to detect rhetorical labels of sentences present in a legal document. CRF is designed in a way similar to how humans summarize legal documents. It works well for text segmentation problems and has proven to be much better than existing models present for text segmentation. CRF is an undirected graph-based structure with conditional probabilities of label sequences given an observation sequence. The conditional probabilities of the label sequences depend on the independent features of the observation sequence. Subsequently, researchers have utilised structures better than CRF for decoding legal documents into meaningful subunits, such as Hierarchical BiLSTM and Hier-BiLSTM CRF classifiers, as discussed next.

### 3.2.2 Hierarchical BiLSTM Classifier

Long short-term memory (LSTM) is a type of recurrent neural network, capable of efficiently processing sequences of data points, and thus are very useful in handling long inter-connected texts such as sentences and paragraphs. **Bidirectional LSTM (BiLSTM)** consists of two LSTMs: one taking the input in a forward direction, and the other in a backwards direction. Thus, BiLSTMs effectively increase the amount of information/context available to the algorithm.

The Sent2vec model[25] is used to generate sentence embeddings for every sentence of the legal document which is then fed into a Hierarchical-BiLSTM model. The Hierarchical Bi-LSTM structure extracts the necessary features from the sentence embeddings to classify the sentences into rhetorical labels. Hier-BiLSTM is used to detect the rhetorical labels because they tend to perform better than the LSTM model. BiLSTM model determines the features from the sentence embeddings more effectively than LSTM model.

### 3.2.3 Hierarchical BiLSTM-CRF Classifier

The embeddings generated by the Sent2vec model[26] are fed into a Hierarchical-BiLSTM model[27], to extract the necessary features from the sentence embeddings to classify the sentences into rhetorical labels. Then the features generated by the Hierarchical-BiLSTM model is being put into the

---

[22] Ghosh, S. and Wyner, A., 2019, December. Identification of rhetorical roles of sentences in Indian legal judgments. In *Legal Knowledge and Information Systems: JURIX 2019: The Thirty-second Annual Conference* (Vol. 322, p. 3). IOS Press.

[23] Saravanan M., Ontology-Based Retrieval and Automatic Summarization of Legal Judgments. https://www.cse.iitm.ac.in/~ravi/papers/Saravanan_thesis.pdf.

[24] Saravanan, M., Ravindran, B. and Raman, S., 2008. Automatic identification of rhetorical roles using conditional random fields for legal document summarization. In *Proceedings of the Third International Joint Conference on Natural Language Processing: Volume-I*.

[25] Moghadasi, M.N. and Zhuang, Y., 2020, December. Sent2vec: A new sentence embedding representation with sentimental semantic. In *2020 IEEE International Conference on Big Data (Big Data)* (pp. 4672-4680). IEEE.

[26] Moghadasi, M.N. and Zhuang, Y., 2020, December. Sent2vec: A new sentence embedding representation with sentimental semantic. In *2020 IEEE International Conference on Big Data (Big Data)* (pp. 4672-4680). IEEE.

[27] Ghosh, S. and Wyner, A., 2019, December. Identification of rhetorical roles of sentences in Indian legal judgments. In *Legal Knowledge and Information Systems: JURIX 2019: The Thirty-second Annual Conference* (Vol. 322, p. 3). IOS Press.

Conditional Random Field [28, 39] for better understanding of rhetorical labels in terms of sequential order. Some rhetorical labels tend to occur one after another in a legal document which is well captured by CRF.

The models have given good results in terms of detection of rhetorical labels. Researchers have seen that detecting rhetorical labels by using deep learning models has performed better as compared to handcrafted features. The problem with handcrafted features is that creation of handcrafted features requires legal expertise. But again, there are issues with deep models where interpretability of these deep models can be questioned. Interpretability of deep models is always a challenging task and here a similar situation arises, though the final results of the deep models are good.

## 4. Legal Document Summarization

Summarization is typically used to get an extract of a long document or a collection of documents[28], which can help a reader save both time and effort. Likewise, summarizing legal documents is necessary to make legal information easily accessible both for lawyers as well as the common man.[29] Average length of an Indian Supreme Court Case is generally 4,500 words. Some of the important landmark cases often span over hundreds of pages, e.g., https://indiankanoon.org/doc/257876/. Some instances were reported where even the judges from one court were not able to comprehend the judgments from another court[30]. Legal document summarization can be useful in many such scenarios. A summarized legal document can be easily understood by the legal practitioners. Furthermore, it can help other tasks like legal search, where searching on the summarized data can give faster results compared to using the original documents.

### 4.1 Automatic summarization of legal documents

Automatically summarizing legal documents is a challenging task given the volumes of legal case documents available across countries. Various domain independent methods are there for legal document summarization which do not consider the domain specific information while creating the summary thereby creating unbalanced summaries. So, designing summarizers which are domain specific is important because this would help us to create more balanced summaries of legal documents thereby encoding proper positional information as well as rhetorical information from the original legal document thereby creating better legal document summaries.

Existing Legal-IR systems require legal experts to create manually curated summaries for legal documents which is a costly process and involves a lot of manual work. Automatically generating legal summaries can help reduce both the cost and time. For designing proper legal summarizers, we need to have legal guidelines from legal experts about what to include in a legal document

---

summary.[31] There have been legal guidelines for UK, US court cases. Similar attempts have also been made for Indian court cases.

**4.2 Many shades of summarization algorithms**

Researchers have primarily focussed on extractive summarization algorithms to generate suitable summaries of legal documents. The methods tried by the researchers can be divided into four categories:

I. Unsupervised domain independent: This type of summarization algorithms has not been trained on legal documents, as they were not designed exclusively for the legal domain. These algorithms are general purpose summarization algorithms which have not been fine-tuned with any legal data.
II. Supervised domain independent: Supervised domain independent summarization algorithms have been trained on legal documents, although these algorithms were not designed for the legal domain. They are general purpose summarization algorithms fine-tuned with legal data.
III. Unsupervised domain specific: These summarization algorithms have been designed exclusively for the legal domain by taking into consideration legal guidelines and legal parameters. However, they do not need to be trained on legal documents.
IV. Supervised domain specific: Supervised domain specific algorithms are trained on legal documents to finetune the model for legal data. They are designed for the legal domain keeping into consideration the legal guidelines and legal parameters.

Next, we discuss several algorithms belonging to different types in detail.

**4.2.1 Unsupervised domain independent methods**

Unsupervised Domain independent methods are those summarization methods which are not designed specifically for the legal domain and can work in the absence of training data. Now we discuss some unsupervised domain independent methods:

- **Lexrank**: Lexrank[32] is a stochastic graph-based method for finding sentence salience in a text document. Text documents consist of a sequential number of sentences arranged in a particular order. Lexrank creates an adjacency matrix where the values in the matrix are the intra-sentence cosine similarity values. This method is a general-purpose text summarization technique proposed in 2004 and outperforms many centroid based and degree based methods of text summarization. This method is not specific to the legal domain. The method is based on the concept of eigen-vector centrality. On computing degree centrality, we have treated each edge of a graph as an equal vote in determining the sentence importance but here we give different weightage to every edge of the graph while computing sentence importance.

- **Luhn Summarizer**: Luhn algorithm[33] is an algorithm based on the concept of TF-IDF vector. It selects the words of higher importance based on frequency of the words. The words present at the beginning of the document are given higher importance. Luhn summarizer is an unsupervised domain independent method for legal document summarization. This method is designed for general purpose text summarization and hence could also be used for legal document summarization.

---

[31] Bhattacharya, P., Poddar, S., Rudra, K., Ghosh, K. and Ghosh, S., 2021, June. Incorporating domain knowledge for extractive summarization of legal case documents. In *Proceedings of the Eighteenth International Conference on Artificial Intelligence and Law* (pp. 22-31).

[32] Erkan, G. and Radev, D.R., 2004. Lexrank: Graph-based lexical centrality as salience in text summarization. *Journal of artificial intelligence research*, *22*, pp.457-479.

[33] Nenkova, A., & McKeown, K. (2011). Automatic summarization. *Foundations and Trends in Information Retrieval*, *5*(2–3), 103-233.

- **LSA Summarizer**: This is an algebraic-statistical method[34] to determine the inherent word and sentence structures present in a document which is important for determining sentence salience. LSA summarizer does not directly work on the sentence features but on the inherent features present in the underlying structures. LSA summarizer works on the principle of dimensionality reduction. It helps us in understanding what topic is being encoded. It also helps us in studying word association by looking into the text dataset in depth. The raw text data is converted into document-term matrix which is thereby converted to singular value decomposition which finally gives us the topic encoded data.

- **Reduction Summarizer**: It is a graph-based unsupervised text summarization technique[35] which condenses a large graph into smaller components to determine the most essential components of the graph and find the sentences of highest importance in a text document. This is a technique which is not domain specific but acts well in determining the summary of a text document. As this is a graph-based technique, our focus is to determine the importance of a vertex in the graph thereby selecting the most important vertices which thereby helps to select the most important sentences in the entire graph.

- **DSDR:** Document summarization based on data reconstruction (DSDR) is an unsupervised legal domain independent technique to summarise legal documents. From the original legal text first stop word removal and stemming is performed to create a weighted term frequency vector for every sentence. The sentences will be known as the candidate set. Now for any sentence in the document, DSDR[36] will select the related sentences of that sentence from the candidate set and will construct a reconstruction function based on that. Now for the sentences in the original legal document, we pick up those sentences which are best representatives of the original document thereby minimising the reconstruction error.

**4.2.2 Supervised domain independent methods**

Supervised domain independent methods need legal data for training the models; however, the summarization methods are not designed specifically for the legal domain. Next, we discuss some supervised domain independent methods:

- **SummaRunner**: SummaRunner[37] refers to a class of neural network based models which is a supervised generic model that can be used for legal document summarization. There are multiple variations of the summa-runner model that can be used to summarise legal documents based on the application needs. The model has 3 variations namely Attn_RNN, RNN_RNN, CNN_RNN. The Attn_RNN is passed through multiple Bi-GRU attention layers one after another. The RNN_RNN is passed through multiple Bi-GRU attention layers and pooling layers alternatively stacked after one another. The CNN_RNN is passed through multiple CNN layers followed by pooling layers and BI-GRU layers stacked after one another.

---

[34] Yeh, J.Y., Ke, H.R., Yang, W.P. and Meng, I.H., 2005. Text summarization using a trainable summarizer and latent semantic analysis. *Information processing & management*, *41*(1), pp.75-95.

[35] Moawad, I.F. and Aref, M., 2012, November. Semantic graph reduction approach for abstractive Text Summarization. In *2012 Seventh International Conference on Computer Engineering & Systems (ICCES)* (pp. 132-138). IEEE.

[36] He, Z., Chen, C., Bu, J., Wang, C., Zhang, L., Cai, D. and He, X., 2012, July. Document summarization based on data reconstruction. In the Twenty-sixth *AAAI conference on artificial intelligence*.

[37] Nallapati, R., Zhai, F. and Zhou, B., 2017, February. Summarunner: A recurrent neural network based sequence model for extractive summarization of documents. In *Thirty-first AAAI conference on artificial intelligence*.

- **BERTSUM**: This is an effective supervised domain independent method for which this model is trained on CNN and DailyNews article dataset to summarize news content. BERT has been useful in many challenging NLP tasks. BERTSUM[38] can summarize a text corpus. This model has fine tuning layers which allows the model to add legal document context to the model. Moreover, the distilled BERT based model for summarization has similar performance to BERT except for the fact that distilled BERT model takes less space and time. The main difference between BERT and BERTSUM is the addition of the data with symbols which point at the beginning and end of a sentence, so the model can learn sentence representations.

### 4.2.3 Unsupervised domain specific methods

Unsupervised domain specific summarization methods are designed specifically for the legal domain, and they can work without any prior training. Some unsupervised domain specific methods are listed below:

- **LetSum**- LetSum[39] is an unsupervised domain specific algorithm which is primarily designed for US court cases. The algorithm divides a legal document into four parts namely Introduction, Context, Judicial Analysis and Conclusion. The algorithm breaks a legal document into four parts namely Introduction, Context, Judicial Analysis and Conclusion taking 10%, 25%, 60%, 5% from the individual parts of the legal document thereby comprising the final summary. Letsum could also be used for Indian legal data by dividing the data into four parts mentioned earlier. The method is generalizable to the court cases of other jurisdictions. This method looks at the thematic segments of the entire legal document to form the final summary.

- **Case Summarizer**- Case summarizer[40] is a domain specific method which is an automatic tool designed for summarization of legal documents. The method uses a word frequency-based approach coupled with domain information to form a summary of a legal document. The method is generalizable to the court cases of other countries. Summaries are being provided with informative interfaces with abbreviations, relevant heat maps and relevant flexible controls. The method is being evaluated by using ROUGE[41] scores (a widely used metric for evaluating the summary quality) and other human based evaluation systems. Case summarizer acts by performing pre-processing, and then performing TF-IDF based analysis to get the final list of ranked sentences. This is a domain specific algorithm designed for the legal domain.

---

[38] Liu, Y., 2019. Fine-tune BERT for extractive summarization. *arXiv preprint arXiv:1903.10318*.
[39] Farzindar, A. and Lapalme, G., 2004. LetSum, an automatic Legal Text Summarizing System. In *the Seventeenth Annual Conference* on *Legal Knowledge and Information Systems* (Vol. 120, p. 11). IOS Press.
[40] Polsley, S., Jhunjhunwala, P. and Huang, R., 2016, December. Casesummarizer: a system for automated summarization of legal texts. In *Proceedings of COLING 2016, the 26th international conference on Computational Linguistics: System Demonstrations* (pp. 258-262).

- **MMR Algorithm**- MMR[42] is a Min-Min roughness algorithm designed based on Rough Set Theory to handle the uncertainty in the clustering process. While many clustering algorithms exist to cluster legal documents, their implementations are challenging because the data is categorical in nature. MMR is a legal domain specific algorithm designed to be applicable on the legal domain.

- **DELSUMM**- DELSUMM[43] is an unsupervised legal domain specific algorithm designed to summarize legal documents created by using the Integer Linear programming approach. DELSUMM is primarily designed for the Indian Legal data and tested on Indian Supreme court data. The model has shown to perform better than most models in terms of Entire document-wise ROUGE[44] scores and in terms of rhetorical role-wise ROUGE scores. This is an extractive summarization-based approach designed primarily to maximize the Integer Linear Programming (ILP) objective function, which is primarily used to summarize news content and social media posts and is now used to summarize Indian legal data. The method is based on two parameters: informativeness of a sentence, and content words. The method gives different importance to different segments.

- **KMM**- KMM[45] stands for K-mixture model and this is an Unsupervised domain dependent approach to create summary of a legal document by selecting important sentences from a legal document. KMM model is used to detect which sentences to include in a legal document summary and which sentences to exclude. After using this model there is a post summarization step which decides which portion of the ranked list to include in the summary. This method does not use rhetorical roles to create the summary.

**4.2.4 Supervised domain specific methods**

Supervised Domain specific methods are designed specifically for the legal domain. They also need the models to be trained on legal data. We discuss one supervised domain specific method below.

---

- **Chinese Gist-** Chinese Gist[46] is a supervised legal document summarization approach which has been applied on Chinese Supreme court cases and has provided very good results. Various machine learning methods such as gradient boosting, multilayer perceptrons, and LSTM are taken into consideration considering various linguistic, legal and statistical information. Producing a gist of Chinese court cases is a challenging task due to the huge number of Chinese court cases. Therefore, it becomes necessary to create an automatic summarizer which can incorporate the domain expertise of Chinese legal system. The method can be generalised to the legal documents of other jurisdictions as well.

**4.3 Domain specific vs. domain independent algorithms**

Domain specific summarization algorithms encode legal information, effectively creating better legal document summaries compared to domain independent methods. To design automated summarizers, we need to have legal guidelines from legal experts about what to include in a legal summary. There have been such legal guidelines for the UK and US court cases, as well as for the Indian court cases.

For example, the following set of guidelines were used in DELSUMM[47]:
  I. The order of importance of rhetorical roles are Final Judgement > Issue > Facts > Statute, Precedent, Ratio > Argument.
  II. The final Judgement and the Issue are extremely important to form the summary for a legal document. Since these portions are present in smaller proportions these sentences should be included fully into the summary of the legal documents.
  III. The sentence appearing in the various rhetorical segments should be included as follows, Facts- Sentences appearing at the beginning of the legal document, Statute- Sentences that contain citations to an act, Precedent- Sentences which refer to prior cases, Ratio of the decision- Sentences which appear towards the ending of a document.
  IV. The final summary must include sentences containing important portions of the case.

Other domain specific legal algorithms include Case Summarizer, MMR algorithm, LetSum, KMM and Chinese Gist which are designed for specific jurisdictions and can also be generalized to other jurisdictions. Algorithms designed for one jurisdiction can work in other jurisdictions by understanding the underlying principles of the particular algorithm and tuning the data in terms of that algorithm.

**4.4 Future of legal document summarization**

Designing effective legal summarization algorithms requires legal domain knowledge. Building a summarization system which can act like a legal expert, in terms of creating legal document summaries, is thus extremely challenging. Researchers have analysed various summarization algorithms and legal expert generated summaries [18,19], and tried to notice the differences and similarities between the algorithms and the experts. This gives an effective view of how to design legal summarization algorithms such that they are closer to the summary created by legal experts.

---

[46] Liu, C.L. and Chen, K.C., 2019, June. Extracting the gist of Chinese judgments of the supreme court. In *proceedings of the seventeenth international conference on artificial intelligence and law* (pp. 73-82).

[47] Bhattacharya, P., Poddar, S., Rudra, K., Ghosh, K. and Ghosh, S., 2021, June. Incorporating domain knowledge for extractive summarization of legal case documents. In *Proceedings of the Eighteenth International Conference on Artificial Intelligence and Law* (pp. 22-31).

There are two primary pointers needed to design an effective summarization algorithm: (i) the proportion of rhetorical roles, and (ii) the position of the sentences forming the final summary. The proportion of different rhetorical roles to be present in the final summary is quite significant in ensuring a good quality summary. Also creating a balanced summary in terms of the position of the sentences[48] chosen from the legal document is also important. Comparing the segment wise ROUGE[49] scores with the rhetorical distribution of the algorithms can suggest the similarities between the algorithmically generated summaries and expert written ones. In future, researchers may try creating ensembled summarization approaches, which take into account and combine the goodness of every summarization algorithm. For example, the summary may include the sentences which are selected by most algorithms thereby indicating their importance for the summarization task.

## 5. Conclusion

In this chapter, we focused on how Artificial Intelligence (AI) has helped in legal data mining, through the collaboration of legal experts and AI researchers and practitioners, bringing in enhancements in the field of AI and Law. We discussed in detail about legal ontologies which can serve as storehouses for legal information. We also discussed two data mining tasks: rhetorical role detection and summarization of case documents. These two tasks are not fully independent – we described how effective legal summarization algorithms can be designed with the help of rhetorical roles. Beyond these two tasks, there are other data mining tasks which have been greatly benefited by the introduction of Artificial Intelligence: Precedence Retrieval[50], Legal Judgement Prediction[51], Statute Detection, Charge Prediction, etc. However, as with any algorithmic systems, there are always concerns about inaccuracies and biases in the AI models being applied in these applications. Hence, the future research in this domain would need to focus on building the models which are transparent and fair to all stakeholders [21, 47, 48, 52]

Similarly, explainability of AI/ML models applied in the legal domain needs adequate emphasis as the legal decisions often have significant consequences for individuals, organizations, and society at large. When legal document models are used to assist or automate legal tasks, it becomes essential to understand how and why these models arrive at their conclusions. Explainability provides transparency and accountability, allowing stakeholders to assess the validity and reliability of the model's outputs. There are several challenges associated with achieving explainability in models trained for legal tasks. One of the primary challenges is the complexity of legal language and the intricate rules and principles that govern legal reasoning. Legal documents often contain ambiguous or subjective language, and legal tasks involve nuanced analysis and interpretation. Translating these complexities into a machine learning model in a way that can be explained to humans is a significant challenge.

---

[48] Grenander, M., Dong, Y., Cheung, J. C. K., & Louis, A. (2019, November). Countering the Effects of Lead Bias in News Summarization via Multi-Stage Training and Auxiliary Losses. In *Proceedings of the 2019 Conference on Empirical Methods in Natural Language Processing and the 9th International Joint Conference on Natural Language Processing (EMNLP-IJCNLP)* (pp. 6019-6024)

[49] Lin, C.Y., 2004, July. ROUGE: A package for automatic evaluation of summaries. In *Text summarization branches out* (pp. 74-81).

[50] Debbarma, R., Prawar, P., Chakraborty, A., and Bedathur., S., 2023 June. IITDLI: Legal Case Retrieval Based on Lexical Models. In *Proceedings of the Tenth International Competition on Legal Information Extraction/Entailment* (pp. 40-48)

[51] Strickson, B. and De La Iglesia, B., 2020, March. Legal judgement prediction for uk courts. In *Proceedings of the 3rd International Conference on Information Science and Systems* (pp. 204-209).

[52] Chhikara, G., Ghosh, K., Ghosh, S. and Chakraborty, A., 2023. Fairness for both Readers and Authors: Evaluating Summaries of User Generated Content. In *Proceedings of the 46th International ACM SIGIR Conference on Research and Development in Information Retrieval*.

The next chapter describes the application of AI to a more specific problem, that of using predictive analytics for justice delivery. It discusses how that can enable the system to address justice delivery at scale, and needless to say, highlights certain caveats that are crucial to a fair mechanism for the same.